%% file: main.tex
\documentclass{article}

\PassOptionsToPackage{numbers, compress}{natbib}

 \usepackage[dblblindworkshop, final]{neurips_2026}

\usepackage[utf8]{inputenc} 
\usepackage[T1]{fontenc}    
\usepackage{hyperref}       
\hypersetup{hidelinks}
\usepackage{url}            
\usepackage{booktabs}       
\usepackage{amsfonts}       
\usepackage{amsmath}        
\usepackage{mathtools}      
\usepackage{nicefrac}       
\usepackage{microtype}      
\usepackage[table]{xcolor}  
\usepackage{graphicx}       
\usepackage{multirow}
\usepackage{makecell}
\usepackage{enumitem}
\usepackage{pifont}         

\newcommand{\cmark}{\ding{51}}
\newcommand{\xmark}{\ding{55}}

\newcommand{\us}{FlowCF}
\newcommand{\wachter}{Wachter}
\newcommand{\dice}{DiCE}
\newcommand{\revise}{REVISE}
\newcommand{\cchvae}{CCHVAE}
\newcommand{\tabcf}{TABCF}
\newcommand{\dicoflex}{DiCoFlex}

\newcommand{\adult}{Adult}
\newcommand{\bank}{Bank marketing}
\newcommand{\default}{Credit default}
\newcommand{\lending}{Lending Club}
\newcommand{\magic}{MAGIC}
\newcommand{\htru}{HTRU2}

\title{\us: Sparse Counterfactual Explanations for Mixed-Type Tabular Data using Flow Matching}
\workshoptitle{Geometric Distributional Deep Learning}

\author{%
  Emmanouil Panagiotou \\
  Freie Universität Berlin \\
  \texttt{emmanouil.panagiotou@fu-berlin.de}
  \And
  Eirini Ntoutsi \\
  University of the Bundeswehr Munich \\
  \texttt{eirini.ntoutsi@unibw.de}
}

\begin{document}

\maketitle

\input{sections/abstract.tex}
\input{sections/introduction.tex}

\input{sections/relatedwork.tex}

\input{sections/method.tex}
\input{sections/experiments.tex}
\input{sections/results.tex}
\input{sections/conclusion.tex}

\bibliographystyle{plainnat}
\bibliography{bibliography}



\end{document}

%% file: sections/abstract.tex
\begin{abstract}
In the field of Explainable AI (XAI), counterfactual (CF) explanations interpret a model's decision by suggesting the changes to the input that would lead to a more favourable outcome. To be useful in practice, such an explanation should change few features and change them as little as possible, properties known as sparsity and proximity. We observe that existing methods remain limited in this respect, especially for numerical features, whether they are model-agnostic and amortised, or gradient-based with full access to the model. In this paper, we propose \us, a model-agnostic generative method that frames CF generation as \emph{sparse transport} from the factual to the target class. We solve this transport with flow matching, which we extend to mixed feature types with a novel mixed flow operator, and exploit the resulting geometry to optimise for sparsity through a gating network that minimises the number of features the transport changes. Extensive experiments on six benchmark datasets demonstrate that \us\ produces the best numerical sparsity and proximity, changing $29\%$ of the numerical features where the best baseline changes $89\%$, at $70\%$ smaller displacement, while remaining comparable on the other desiderata.
\end{abstract}

%% file: sections/introduction.tex
\section{Introduction}
\label{sec:intro}

Given a model's decision on an instance, a counterfactual (CF) explanation suggests the changes to the input that would lead to a more favourable outcome. This answers a what-if question about that instance, which makes CFs directly actionable, and a common choice in domains where decisions affect individuals, such as credit, hiring, and medicine, where the data is typically tabular, with both numerical and categorical features \cite{tabcf}.

A useful CF has to satisfy several, partly conflicting, properties, which the literature refers to as desiderata \cite{verma2024counterfactual}. The most common are \emph{validity} (the prediction changes), \emph{proximity} (the CF stays close to the factual instance), \emph{sparsity} (few features change), and \emph{actionability} (it respects what the user is able to change), while many more have been proposed, such as \emph{plausibility}, \emph{diversity}, and \emph{amortised inference}, with each method typically targeting a different subset of them. Notably, sparsity and proximity determine how easy a CF is to understand and act on, since explanations are more effective when they involve few features \cite{miller2019explanation}, and smaller changes are more realistic to implement \cite{verma2024counterfactual,dice}.

Existing methods approach CF explanation generation in different ways and assume different access to the model they explain, but all solve the same underlying task: moving an instance from one class (decision) to a targeted other. Optimisation-based methods such as \dice\ \cite{dice} and \tabcf\ \cite{tabcf} search for counterfactuals using gradient descent, in the input space or latent space, which requires access to gradients through the model and one optimisation run per explanation. Generative methods such as \cchvae\ \cite{cchvae} and \dicoflex\ \cite{dicoflex} instead learn the structure of the data with a variational autoencoder or a normalising flow, and can amortise generation into a single forward pass.

\begin{figure}[t]
  \centering
  \includegraphics[width=0.7\linewidth]{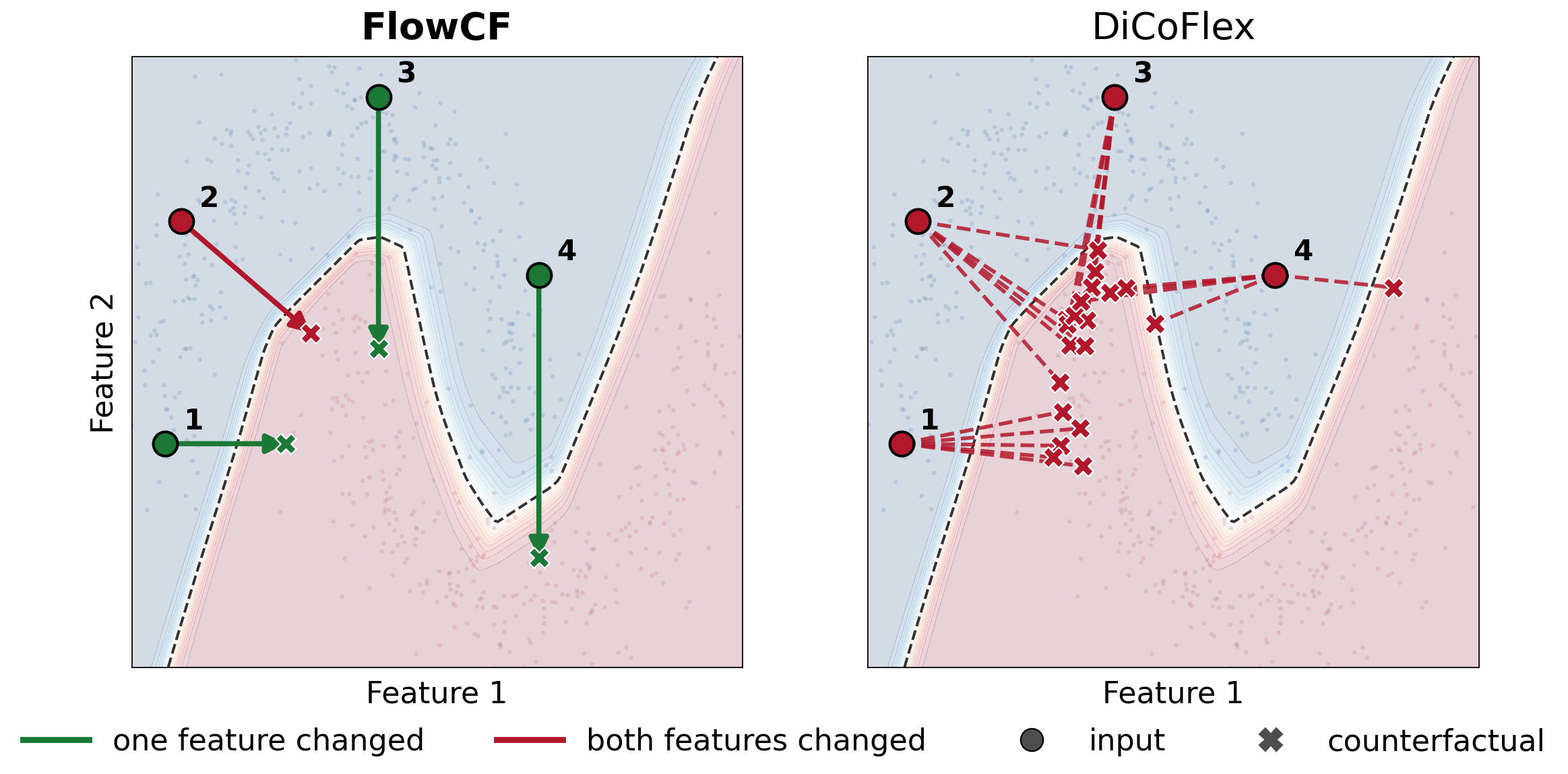}
\caption{Counterfactuals generated by our method \us\ vs \dicoflex, the closest baseline method, on the two-moons dataset. \us\ achieves exact numerical sparsity, changing a single feature when possible, while all counterfactuals sampled by \dicoflex\ change both features.}
  \label{fig:moons}
\end{figure}

In this work, we observe that existing methods fall short on sparsity and proximity, especially for numerical features. Exact numerical sparsity is difficult to optimise for by construction, since it is defined as the $L_0$ count of changed features, which is non-differentiable and therefore not directly optimisable by gradient descent \cite{l0reg}. Existing methods approximate it with an $L_1$ term or heuristics such as post-hoc pruning, neither of which leaves a feature exactly unchanged (see Figure~\ref{fig:moons}).
\input{sections/table_access}
To address these limitations, we introduce \us, a generative method that frames CF generation as \emph{sparse transport} from the factual class to the target class. Flow matching is a natural fit for this setting, as it learns a velocity field that moves samples from one continuous distribution to another. Since classical flow matching is defined for continuous data only, we extend it to mixed feature types with a mixed flow operator, following categorical flow maps \cite{catflowmaps}. The operator takes a single step along a straight path, so that each instance is explained by one displacement vector over its features, and we exploit this geometry to optimise for sparsity. Sparsity is achieved via a gating network that minimises the number of features the transport changes, and proximity by minimising the flow vectors. As summarized in Table~\ref{tab:access}, \us\ is the only method that is simultaneously model-agnostic and amortised, requires no model access at explanation time, and achieves exact numerical sparsity. We perform a quantitative evaluation against state-of-the-art baselines on six binary classification datasets (Section~\ref{sec:results}), showing that \us\ produces the sparsest and most proximal numerical counterfactuals on every dataset, while remaining comparable on the other desiderata.

Our contributions are: i) identifying the limitation of existing CF methods in achieving exact numerical sparsity; ii) introducing \us, a novel method using a gated, mixed flow-matching transport for generating sparse and proximal counterfactual explanations. The code for reproducing our experiments is included in the supplementary material.

%% file: sections/table_access.tex
\begin{table}[t]
  \caption{Desired properties of existing counterfactual methods compared to \us.}
  \label{tab:access}
  \setlength{\tabcolsep}{6pt}
  \centering
  \begin{tabular}{l c c c c}
    \toprule
    \midrule
     & \makecell{model-\\agnostic} & amortised & \makecell{exact numerical\\sparsity} & \makecell{no black-box queries\\at inference}\\
    \midrule
     \wachter   & \xmark & \xmark & \xmark & \xmark\\
     \dice      & \xmark & \xmark & \xmark & \xmark\\
     \tabcf     & \xmark & \xmark & \xmark & \xmark\\
     \revise    & \xmark & \xmark & \xmark & \xmark\\
     \cchvae    & \cmark & \xmark & \xmark & \xmark\\
     \dicoflex  & \cmark & \cmark & \xmark & \xmark\\
     \textbf{\us} & \cmark & \cmark & \cmark & \cmark\\
  \bottomrule
  \bottomrule
  \end{tabular}
\end{table}

%% file: sections/relatedwork.tex
\section{Related work}
\label{sec:related}
The literature on counterfactual explanations is extensive \cite{verma2024counterfactual,carla}. We cover the works closest to ours: counterfactual generation for tabular data, and flow-based transport.

\noindent{\textbf{Counterfactual explanations for tabular data.}} Wachter et al. \cite{wachter} first framed CF search as an optimisation problem in input space, minimising the distance to the original instance subject to a label flip. This inspired methods that assume a differentiable black box and take gradient steps, either in input space like \dice\ \cite{dice}, or in the latent space of a generative model like \revise\ \cite{revise} and \tabcf\ \cite{tabcf}. Sparsity can enter their objective, through an $L_1$ term or a post-hoc pruning step, but at the cost of a separate optimisation run per instance and gradient access to the model being explained. A second family avoids per-instance optimisation. \cchvae\ \cite{cchvae} performs a growing-sphere search in the latent space of a conditional VAE, and \dicoflex\ \cite{dicoflex} trains a conditional normalising flow that emits multiple diverse counterfactuals in a single forward pass, with constraint control applied at inference time. Both are model-agnostic, requiring no gradients through the black box, and \dicoflex\ additionally amortises generation. Both methods, however, produce dense counterfactuals, especially for numerical features. \dicoflex\ targets sparsity only through its nearest-neighbour selection, preferring target-class neighbours that differ in fewer features, which encourages small changes but never exactly freezes a feature (see Figure~\ref{fig:moons}).

\noindent{\textbf{Flow-based generation.}} Flow-based methods are generative models that learn to map one, usually continuous, distribution onto another. Normalising flows \cite{rezende2015variational} do so through an invertible transformation of the base density, giving exact likelihoods at the cost of an architecture restricted to invertible maps with a tractable Jacobian, and of learning a density rather than a movement between distributions. \emph{Flow matching} \cite{flowmatching} lifts both restrictions by prescribing an interpolant between the two distributions and regressing a velocity field along it, removing the need for invertibility and for simulating an ODE during training. Since a convex interpolant is undefined for discrete variables, the formulation has been extended to categorical data through the probability simplex \cite{catflowmaps}. Building on this, rectified flows \cite{rectifiedflow} take the straight-line interpolant between two observed distributions, and show that straightening the paths lets a single Euler step replace integrating the ODE. Normalising flows have been used for counterfactual generation \cite{ceflow,dicoflex}, focusing on plausibility and diversity. We propose a mixed \emph{flow matching} operator that transports between the two class-conditional distributions in a single step, using the resulting geometry to optimise for exact sparsity, proximity, and validity.

%% file: sections/method.tex
\section{\us: sparse flow-matching counterfactual explanations}
\label{sec:method}

Following prior model-agnostic work \cite{dicoflex}, we consider a binary black-box classifier $f:\mathcal{X}\rightarrow y \in \{0,1\}$. We observe only its predicted labels, without access to gradients, model parameters, probabilities, or logits. Each input $x=[x^{\mathrm{num}},x^{\mathrm{cat}}]$ contains $|N|$ numerical and $|C|$ categorical features. Let $\pi_0$ and $\pi_1$ denote the empirical distributions of instances assigned to the factual and target classes, respectively. We define the \emph{sparse transport problem} as learning a map $\mathcal{T}$ that moves an instance $x\sim\pi_0$, with $f(x)=0$, toward $\pi_1$, producing a valid counterfactual $x'=\mathcal{T}(x)$ with $f(x')=1$ while changing as few features as possible.

Flow matching is a natural fit for this problem because it learns a transport between distributions from paired samples. Like other generative counterfactual methods \cite{dicoflex}, it amortises generation and avoids a separate optimisation for each counterfactual. However, standard flow matching does not account for CF desiderata like sparsity and proximity, and therefore produces dense displacements.

To address this limitation, we introduce \us. Our method is comprised of two models and is trained in three steps. The first model is a mixed flow operator $F_\theta$, designed to support mixed-type data and move factual instances toward the target class. The second model is a gate operator $G_\psi$ that determines which feature changes are applied to each instance. At inference, the gates are binarised so that inactive features remain exactly unchanged.

\begin{enumerate}[label=\textbf{Step \arabic*.},leftmargin=*,align=left,labelsep=0.5em,itemsep=2pt,topsep=4pt]
\item \textbf{Transport} (Section~\ref{sec:flow}). We train $F_\theta$ on pairs of factual and target-class instances to learn target-directed feature changes.
\item \textbf{Sparsity} (Section~\ref{sec:gate}). We freeze $F_\theta$ and train $G_\psi$ to remove unnecessary feature changes while maintaining validity.
\item \textbf{Joint} (Section~\ref{sec:joint}). We freeze $G_\psi$ and fine-tune $F_\theta$ through the learned gates to improve proximity while maintaining validity and staying close to the transport learned in Step~1.
\end{enumerate}

\begin{figure}[htbp]
\centering
\includegraphics[width=\linewidth]{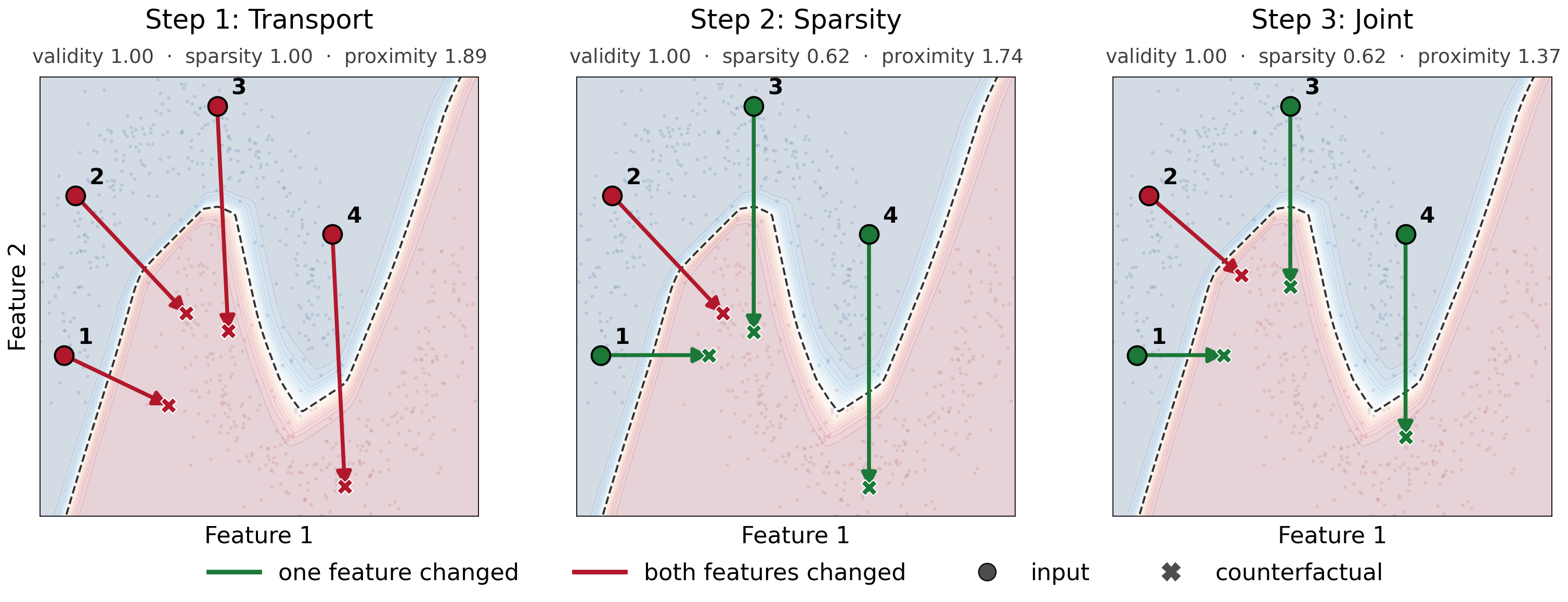}
\caption{\us's three training steps on samples from the 2D two-moons dataset. Green marks a learned displacement that changed one feature, red both. Step 1 produces valid CFs but changes both features. Step 2 achieves better sparsity making three of the four CFs axis-aligned. Step 3 further improves proximity while preserving validity and sparsity.}
\label{fig:threesteps}
\end{figure}

Figure~\ref{fig:threesteps} illustrates the effect of each training step on the same four factual instances.

\subsection{Step 1: Transport}\label{sec:flow}
The first step learns a transport from $\pi_0$ to $\pi_1$. Since a factual instance can be paired with several target-class instances, this pairing directly affects the proximity and sparsity of the resulting counterfactual. We pair each $x_0$ with its $K$ nearest neighbours in the target class, using $L_1$ distance on the numerical block and Hamming distance on the categorical block, following \cite{dicoflex}. This keeps the transport short and ensures that each path ends at an observed target-class instance.

We learn the transport between these pairs with flow matching \cite{rectifiedflow,flowmatching}. For the numerical block, we take the interpolant \  $ x^{\mathrm{num}}_t = (1-t)\,x^{\mathrm{num}}_0 + t\,x^{\mathrm{num}}_1, \ t \sim \mathcal{U}[0,1] \ $ and train a velocity field $v_\theta$ to predict the displacement from the intermediate point. However, this does not extend to categorical features, since a convex combination of two one-hot vectors is not a valid category. Following the \emph{categorical flow} method \cite{catflowmaps}, we represent each categorical column in the probability simplex and predict an endpoint distribution $\mu_\theta$ rather than a velocity. The two parameterisations are equivalent along a straight-line interpolant \cite{catflowmaps}, while the endpoint form keeps categorical predictions admissible.

Our mixed flow operator $\mathcal{F}_\theta$ supports both output feature types present in the data. It \emph{moves} the numerical block by a displacement, and \emph{replaces} the categorical block with a predicted endpoint:
\begin{equation}
\mathcal{F}_\theta(x,t): \qquad
x^{\mathrm{num}} \;\longmapsto\; x^{\mathrm{num}} + \underbrace{v_\theta(x,t)}_{\mathclap{\text{numerical displacement}}},
\qquad
x^{\mathrm{cat}} \;\longmapsto\; \underbrace{\mu_\theta(x,t)}_{\mathclap{\text{categorical endpoint}}},
\label{eq:flowop}
\end{equation}
We optimise $\mathcal{F}_\theta$ using the following loss:
\begin{align}
\mathcal{L}_{\text{flow}}
={}& \mathbb{E}\Big[
    \big\|v_\theta(x_t,t)
    - (x^{\mathrm{num}}_1-x^{\mathrm{num}}_0)\big\|_2^2
    + \lambda_{\mathrm{cat}}\,
    \mathrm{CE}\big(\mu_\theta(x_t,t),x^{\mathrm{cat}}_1\big)
    \Big]
    + \lambda_{\mathrm{val}}\,
    \mathcal{L}_{\text{val}}\big(\mathcal{F}_\theta(x)\big),
    \notag\\
&\text{where}\quad
\mathcal{L}_{\text{val}}\big(\mathcal{F}_\theta(x)\big)
= \mathbb{E}\Big[
    \operatorname{hinge\_loss}
    \big(S(\mathcal{F}_\theta(x)),y=1\big)
    \Big].
\end{align}
The first term learns the numerical and categorical mappings. The numerical velocity is trained by squared error, while $\mathrm{CE}$ is the cross-entropy between the predicted and paired target categories, summed over categorical columns using the mean-field factorisation of \cite{catflowmaps}. The weight $\lambda_{\mathrm{cat}}$ balances the two feature types.

The second term provides a direct validity signal. Since $f$ is a black-box classifier, we train a surrogate $S$ to reproduce its predictions on $\mathcal{D}$ and freeze it. We then define $\mathcal{L}_{\mathrm{val}}$ using the hinge loss of \cite{dice}, with $\lambda_{\mathrm{val}}$ controlling its contribution. The hinge loss stops penalising a counterfactual once it reaches the target margin, avoiding unnecessary movement deeper into the target class.

We generate using a single Euler step from $t=0$, i.e. a one-step map $\mathcal{F}_\theta(x):=\mathcal{F}_\theta(x,0)$. We find this sufficient for tabular data, and consistent with prior work showing that rectified flows can be distilled into accurate one-step models for fast inference even in high-dimensional domains \cite{rectifiedflow}.

\subsection{Step 2: Sparsity}
\label{sec:gate}
The learned flow of Step 1 changes every feature, without taking the \emph{sparsity} desideratum into account. To address this, we freeze the transport $\mathcal{F}_\theta$ and introduce a gating network, trained to zero out as many feature changes as possible while preserving validity. What we want is a binary mask over features, selecting the ones the flow may change, but such a mask cannot be directly learned by gradient descent, as neither the mask nor the $L_0$ count is differentiable. Following $L_0$-style gating \cite{l0reg,l0gated}, we relax it to a continuous gate $g$, produced by a network $a_\psi$ with one output per feature. Our counterfactual operator $\mathcal{T}_{\theta,\psi}$ then applies the flow to each feature in proportion to its gate:
\begin{equation}
\mathcal{T}_{\theta,\psi}(x) \;=\; \underbrace{(1 - g) \odot x}_{\text{keep the factual}} \;+\; \underbrace{g \odot \mathcal{F}_\theta(x)}_{\text{apply the flow}}, \qquad g \;=\; \mathcal{G}_\psi(x) \;=\; \sigma\!\big(a_\psi(x) / T\big),
\label{eq:gated}
\end{equation}
with $\sigma$ the logistic sigmoid and $T$ a temperature controlling the hardness of the relaxation. We keep the gate deterministic, so that training and inference remain consistent. A gate of $0$ keeps a feature at exactly its factual value and a gate of $1$ applies the flow in full, while intermediate values scale the displacement of a numerical feature or interpolate a categorical one towards the predicted endpoint. Our gate also naturally guarantees the \emph{actionability} desideratum: setting $g_c \equiv 0$ for immutable features strictly enforces the constraint, rather than soft-penalizing it, forcing the transport to reach the target class through actionable features alone.

We train the gate $\mathcal{G}_\psi$ with three terms, one per desideratum, optimising the composed map $\mathcal{T}_{\theta,\psi}$ of Equation~\ref{eq:gated} with the transport $\mathcal{F}_\theta$ frozen:
\begin{equation}
\mathcal{L}_{\text{gate}} \;=\; \underbrace{\mathcal{L}_{\text{val}}(\mathcal{T}_{\theta,\psi}(x))}_{\text{validity}} \;+\; \lambda_{\mathrm{sparse}}\underbrace{\mathcal{L}_{\text{sparse}}}_{\text{sparsity}} \;+\; \lambda_{\mathrm{prox}}\underbrace{\mathcal{L}_{\text{prox}}}_{\text{proximity}},
\label{eq:gateloss}
\end{equation}
\begin{equation}
\text{where}\quad \mathcal{L}_{\text{sparse}} \;=\; \mathbb{E}\Big[\sum_{c \in N \cup C} g_c\Big], \qquad \mathcal{L}_{\text{prox}} \;=\; \mathbb{E}\Big[\sum_{c \in N} |g_c\, v^c_\theta| \;+\; \sum_{c \in C} g_c\, p_c\Big].
\label{eq:gateterms}
\end{equation}
We use the same validity term as in Step 1, now evaluated on the \emph{gated} counterfactual rather than on the raw endpoint of the transport. Gating a feature the label flip depends on is therefore penalised; without this signal, the gate would optimise sparsity and proximity alone, and simply keep the features that change least. The sparsity term is the differentiable surrogate for the $L_0$ count of the sparsity metric in Section~\ref{sec:metrics}, charging for every feature the gate leaves open. Its weight $\lambda_{\mathrm{sparse}}$ controls how sparse the edit is, closing more features as it increases, until validity begins to suffer. The proximity term charges the scale of the displacement identically for both feature types, using the flow's displacement $v^c_\theta$ for numerical features, and for categorical ones the probability mass $p_c$ that the predicted endpoint places outside the original category.

\subsection{Step 3: Joint}
\label{sec:joint}
In the final step we unfreeze the transport $\mathcal{F}_\theta$ and freeze the gate $\mathcal{G}_\psi$, optimising the composed map $\mathcal{T}_{\theta,\psi}$ of Equation~\ref{eq:gated} against the same validity and proximity terms. The flow is therefore re-fit \emph{through} the gate, so that only the features the gate left open can carry the label flip. To avoid unstable training, a third term $\mathcal{L}_{\text{drift}}$ keeps $\mathcal{F}_\theta$ close to its Step 1 solution $\mathcal{F}_{\theta'}$.
\begin{equation}
\mathcal{L}_{\text{joint}} \;=\; \mathcal{L}_{\text{val}}(\mathcal{T}_{\theta,\psi}(x)) \;+\; \lambda_{\mathrm{prox}}\,\mathcal{L}_{\text{prox}} \;+\; \lambda_{\mathrm{drift}}\,\mathcal{L}_{\text{drift}}, \qquad
\mathcal{L}_{\text{drift}} = \mathbb{E}\big[\,\|\mathcal{F}_\theta(x) - \mathcal{F}_{\theta'}(x)\|_2^2\,\big].
\label{eq:jointloss}
\end{equation}
This joint training step improves proximity by shortening the displacement within the features the gate kept open, while preserving sparsity and validity. We visualise the effect in the second and third panels of Figure~\ref{fig:threesteps}.

\textbf{Inference.} Our final counterfactual model is the composed map $\mathcal{T}_{\theta,\psi}$, given by the transport $\mathcal{F}_\theta$ of Step 3 and the gate $\mathcal{G}_\psi$ of Step 2. Explaining an instance requires one forward pass of each: we evaluate $\mathcal{F}_\theta$ and the binarised gate $\mathcal{G}_\psi$, and apply Equation~\ref{eq:gated}. Open categorical features take the arg-max of the predicted endpoint, snapping it back to a vertex of the simplex, while closed features keep the factual value exactly. The black-box model is never called.

%% file: sections/experiments.tex
\section{Experimental evaluation}
\label{sec:experiments}
This section presents the setup, datasets, baselines, and metrics of our quantitative evaluation.

\subsection{Experimental setup}
\label{sec:setup}
Following the literature on CF evaluation \cite{tabcf,carla}, all methods run through a single evaluation pipeline, so that factual selection, black-box training and metric computation are identical across methods. For each dataset we train one shared black box, a single $1000$ hidden layer MLP. Every method explains the same $n = 1000$ previously unseen instances assigned to the undesired class. A valid CF is not guaranteed for every instance \cite{tabcf}, and every metric other than validity is computed over the valid counterfactuals only.

For \us\ we tune the hyperparameters manually and use the resulting configuration for all datasets, with no per-dataset tuning, $\{ d_S = 256 \times 4$, $d_F = 128 \times 3$, $K = 10$, $\lambda_{\mathrm{cat}} = 1.0$, $\lambda_{\mathrm{val}} = 2.0$, $\lambda_{\mathrm{sparse}} = 0.3$, $\lambda_{\mathrm{prox}} = 0.375$, $\lambda_{\mathrm{drift}} = 0.05$, $T = 0.2 \}$, where $d_S$ and $d_F$ denote the width and depth of the surrogate and flow networks. All experiments use a single 16 GB RTX 3080 Ti GPU. \us\ completes training and inference in under 35 seconds per dataset, substantially faster than several optimisation-based baselines.




\subsection{Datasets}
\label{sec:datasets}
We evaluate on six binary classification datasets, listed in Table~\ref{tab:datasets}. Four are mixed-type: \adult, \bank, \default\ and \lending\ \cite{frank2010uci,jagtiani2019roles}, all standard in the tabular counterfactual literature \cite{tabcf,carla}. Two datasets are continuous-only: \magic\ \cite{frank2010uci} and \htru\ \cite{htru2}, with only floating-point features. They are deliberately included to showcase the importance of the exact numerical sparsity \us\ achieves.
\input{sections/table_datasets}

We further perform a qualitative visual experiment by encoding the CelebA-HQ \cite{celebahq} dataset, using VecGAN++ \cite{vecganpp}, which factorises the latent space into six continuous, interpretable feature axes (\texttt{bangs}, \texttt{glasses}, \texttt{hair}, \texttt{young}, \texttt{male}, \texttt{smile}). A counterfactual image is a decoded displacement along these axes. For evaluation, a binary image-space black-box classifier scores the celebrity images for a target attribute (e.g.\ \texttt{male} or \texttt{young}).

\subsection{Baselines}
\label{sec:baselines}

We compare against six methods, using CARLA reference implementations \cite{carla} for the first four and author implementations for the rest, evaluated against the same black box and factual instances.

\noindent\textbf{\wachter} \cite{wachter} minimizes distance to the original instance subject to a label flip via input-space gradient descent, perturbing numerical features only. \textbf{\dice} \cite{dice} optimizes in input space with a regularization term to normalize categorical one-hot blocks and project categories via arg-max, using a post-hoc pass to restore unnecessary numerical edits. \textbf{\revise} \cite{revise} optimizes the same objective in a VAE latent space, decoding at each step to measure input-space distance. \textbf{\cchvae} \cite{cchvae} searches the latent space of a conditional VAE with a growing-sphere procedure until a decoded sample flips the model, avoiding gradients but requiring many queries. \textbf{\tabcf} \cite{tabcf} optimizes in the latent space of a transformer-based VAE using a Gumbel-Softmax detokenizer to reconstruct categorical features without post-hoc rounding. \textbf{\dicoflex} \cite{dicoflex} is the closest baseline to our method, using a conditional normalizing flow to generate multiple candidates in one forward pass (amortized), selecting CFs from up to 100 generated candidates per sample based on validity. While DiCoFlex targets sparsity through low-$p$ neighbour selection, it does so only softly via numerical $\epsilon$-sparsity.


In contrast, \us\ achieves exact numerical sparsity: our hard-thresholding gate optimally freezes features during generation, keeping unedited numerical features identical to the factual instance.

\subsection{Metrics}
\label{sec:metrics}

We evaluate along the desiderata of Section~\ref{sec:intro}, and in line with literature \cite{carla}, computing every metric directly in the input space so that all methods are scored identically.

\noindent{\textbf{Validity.}} The share of explained instances whose counterfactual flips the black box:
\[
\text{Validity }(\uparrow) \;=\; \frac{n_{\mathrm{val}}}{n} \;=\; \frac{1}{n}\sum_{i=1}^{n} \mathbb{I}\big(f(x'_i) = 1\big).
\]

\noindent{\textbf{Sparsity.}} The fraction of features changed, reported separately per feature type:
\begin{align*}
\text{Sparsity Cat }(\downarrow) &= \frac{1}{n_{\mathrm{val}}}\sum_{i=1}^{n_{\mathrm{val}}} \frac{\|x_0^{\mathrm{cat}} - x'^{\,\mathrm{cat}}\|_0}{|C|}, &
\text{Sparsity Num }(\downarrow) &= \frac{1}{n_{\mathrm{val}}}\sum_{i=1}^{n_{\mathrm{val}}} \frac{\|x_0^{\mathrm{num}} - x'^{\,\mathrm{num}}\|_0}{|N|}.
\end{align*}

\noindent{\textbf{Proximity.}} The distance between the counterfactual and the factual instance, measuring how close the explanation remains to the original input rather than how many features it modifies. We standard-normalize the numerical features and average the absolute displacement over $|N|$:
\[
\text{Prox Num }(\downarrow) \;=\; \frac{1}{n_{\mathrm{val}}}\sum_{i=1}^{n_{\mathrm{val}}} \frac{\big\|x_0^{\mathrm{num}} - x'^{\,\mathrm{num}}\big\|_1}{|N|}.
\]

Here, $\|\cdot\|_0$ and $\|\cdot\|_1$ denote the $L_0$ (count) and $L_1$ distance, respectively.

%% file: sections/table_datasets.tex
\begin{table}[htbp]
  \caption{Dataset characteristics.}
  \label{tab:datasets}
  \setlength{\tabcolsep}{6pt}
  \centering
  \begin{tabular}{l c c l}
    \toprule
    \midrule
    \textbf{Dataset} & \makecell{\textbf{Training}\\\textbf{size}} & \makecell{\textbf{\#Features}\\$\mathbf{(Num/Cat)}$} & \textbf{Classification task}\\
    \midrule
     \adult   & 32{,}561 & 4 / 8  & Income above or below \$50K\\
     \bank    & 20{,}002 & 7 / 9  & Subscription to a term deposit\\
     \default & 27{,}000 & 14 / 9 & Default on the next payment\\
     \lending & 50{,}000 & 8 / 4  & Loan fully paid or charged off\\
     \magic   & 9{,}510  & 10 / 0 & Gamma versus hadron air shower\\
     \htru    & 8{,}949  & 8 / 0  & Pulsar versus spurious candidate\\
  \bottomrule
  \bottomrule
  \end{tabular}
\end{table}

%% file: sections/results.tex
\section{Results}
\label{sec:results}

This section presents our quantitative results on all tabular datasets (Section~\ref{sec:results_tabular}), followed by our visual qualitative results on CelebA-HQ (Section~\ref{sec:results_image}).

\subsection{Results for all baselines on all tabular datasets}
\label{sec:results_tabular}

We report the results averaged over all datasets in Table~\ref{tab:summary}, and the per-dataset breakdown in Table~\ref{tab:results}.
\input{sections/table_summary}

On average, \us\ achieves the best numerical sparsity and the best numerical proximity of all methods: we change $29\%$ of the numerical features where the best baseline changes $89\%$, and we change them by $0.226$ against $0.763$, a $70\%$ smaller displacement. We are also essentially the only method that achieves numerical sparsity at all, as every baseline sits between $89\%$ and $100\%$, changing almost every numerical feature of almost every instance. \dicoflex\ is the only competitor that targets numerical sparsity directly, but does so softly, encouraging small changes rather than freezing features, as we do with our gating mechanism. Although we do not outperform on validity and categorical sparsity, \us\ remains competitive on both.

From the per-dataset results of Table~\ref{tab:results} we observe that \us\ is first on numerical sparsity and numerical proximity on all six datasets, with no exceptions. The effect is clearest on the two continuous-only datasets, where every competitor changes every numerical feature of every instance (Sparsity Num $= 100\%$), while \us\ changes only $19\%$ on \magic\ and $41\%$ on \htru. Validity stays above $95\%$ on four of the six datasets, and drops only on \bank\ and \default. Categorical sparsity is likewise comparable to the other amortised and latent-space methods, \tabcf\ and \dicoflex, on every dataset.

\input{sections/table_results}

\subsection{Visual qualitative results on CelebA-HQ}
\label{sec:results_image}

\begin{figure}[t]
  \centering
  \includegraphics[width=0.92\linewidth]{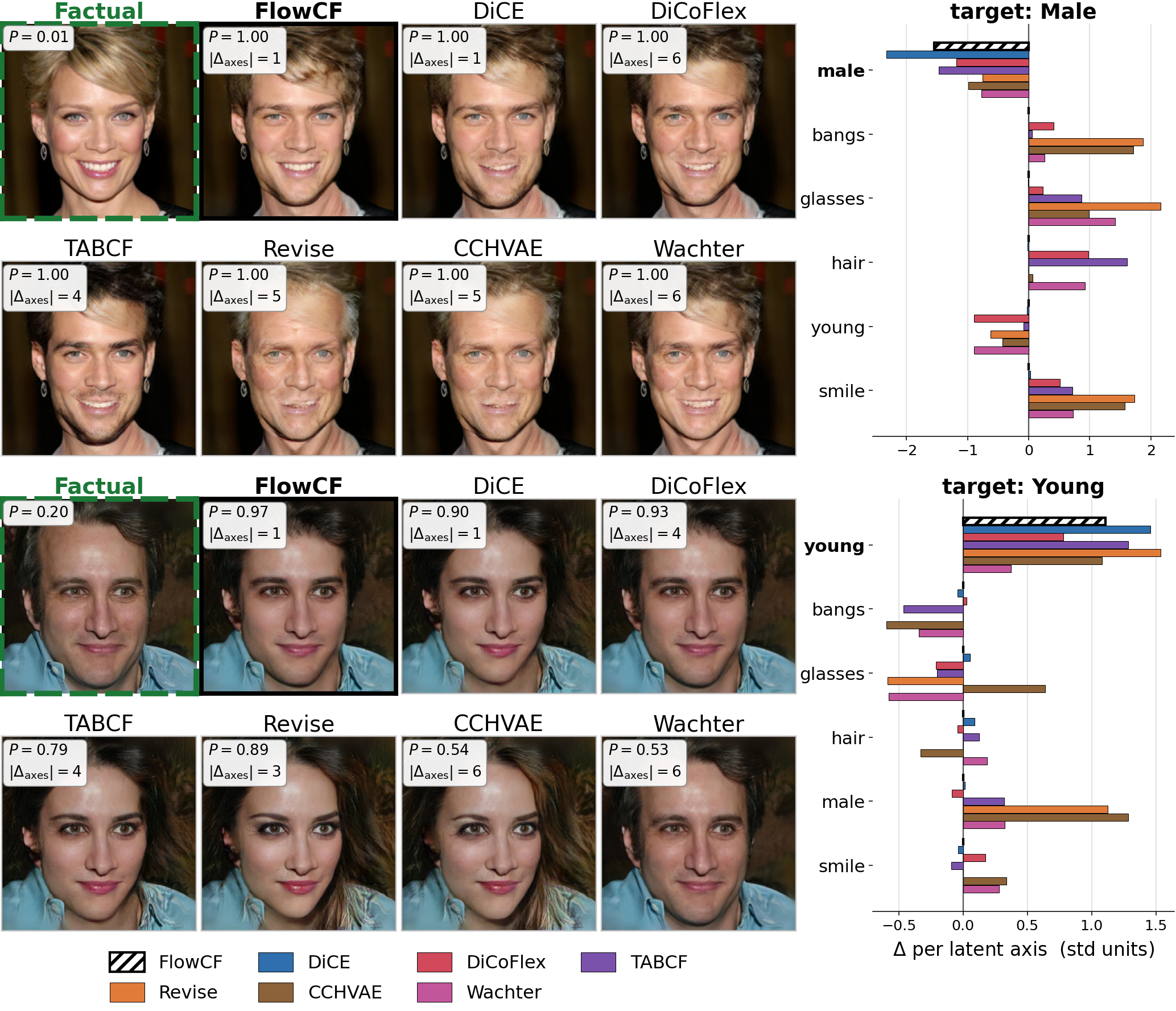}
\caption{Counterfactuals generated by all methods for two CelebA-HQ instances, targeting the \texttt{male} and \texttt{young} attributes. The factual image is marked in green, and each image is annotated with the black-box's probability $P$ of the target class and the number of changed axes $|\Delta_{\text{axes}}|$. The bar plot gives the per-axis displacement in standard deviations, with our method \us\ hatched.}
  \label{fig:vecgan}
\end{figure}

Figure~\ref{fig:vecgan} shows two visual examples from the CelebA-HQ dataset. On the left (in green) is the factual image, followed by the counterfactual images generated by all methods. As described in Section~\ref{sec:datasets}, each method operates on the six-dimensional VecGAN++ \cite{vecganpp} latent features, rather than on the raw image, so a counterfactual is a displacement along the six feature axes, which we report on the right, and the resulting image is obtained via decoding \cite{vecganpp}. For the first instance the binary classification target was chosen to be the \texttt{male} attribute and for the second the \texttt{young} attribute. Each face image is annotated with the probability the black-box assigns to the target class, and with the number of axes (features) the method changed.

Every method flips the black-box decision on both targets, creating valid counterfactuals. \us\ (hatched black bar) achieves sparsity by changing a single axis, and that axis is the correct one, identified without any prior knowledge of which feature the classifier depends on for each target setting. \dice\ (blue bar) also achieves sparsity, but with worse proximity by making larger changes along that axis. This difference is visually apparent as our generated counterfactuals are more faithful to the original factual images.

The bar plot shows what the remaining methods change instead, and the effect is equally visible in the decoded images. On the first instance, \tabcf\ turns the hair from blonde to black, while \revise\ and \cchvae\ push the \texttt{young} axis and make the subject look older. Both are attributes correlated with the \texttt{male} target in this dataset, where blonde hair is rare among male faces and most males are older. The second instance shows the same effect in the opposite direction, where several methods push the \texttt{male} axis strongly towards the female direction, since youth is correlated with females in this dataset.

%% file: sections/table_summary.tex
\begin{table}[htbp]
  \caption{Results averaged over all datasets. \textbf{Top 2} counts best or runner-up finishes in Table~\ref{tab:results}.}
  \label{tab:summary}
  \setlength{\tabcolsep}{2pt} 
  \centering

  \begingroup
  \begin{tabular}{l c c c c c}
    \toprule
    \midrule
     & \textbf{Validity} (\%) $\uparrow$ & \multicolumn{2}{c}{\textbf{Sparsity}} (\%) $\downarrow$ & \textbf{Proximity} $\downarrow$ & \textbf{Top 2} $\uparrow$\\
     & & \textbf{Cat} & \textbf{Num} & \textbf{Num} & \\
     \midrule

     \wachter & 0.863 & - & 1.000 & 1.113 & 0\\
     \dice & \textbf{1.000} & \textbf{0.145} & 0.893 & 0.957 & \textbf{15}\\
     \revise & 0.717 & 0.172 & 0.990 & 0.819 & 6\\
     \cchvae & 0.900 & 0.196 & 1.000 & 0.763 & 6\\
     \tabcf & 0.926 & 0.458 & 0.898 & 1.164 & 6\\
     \dicoflex & \textbf{1.000} & 0.411 & 0.953 & 0.824 & 6\\
     \textbf{\us\ (us)} & 0.939 & 0.394 & \textbf{0.287} & \textbf{0.226} & \textbf{15}\\

  \bottomrule
  \bottomrule
  \end{tabular}
  \endgroup
\end{table}

%% file: sections/table_results.tex
\begin{table}[htbp]
\caption{Detailed results on each dataset. \textbf{Bold} is best, \underline{underlined} second best. Categorical sparsity does not apply to the continuous-only datasets \magic\ and \htru.}
  \label{tab:results}
  \setlength{\tabcolsep}{1.2pt} 
  \centering

  \begingroup\small
  \begin{tabular}{l c c c c @{\hspace{1.6em}} c c c c}
    \toprule
    \midrule
     & \multicolumn{4}{c}{} & \multicolumn{4}{c}{}\\[-2.2ex]
     & \textbf{Validity} (\%) $\uparrow$ & \multicolumn{2}{c}{\textbf{Sparsity}} (\%) $\downarrow$ & \textbf{Proximity} $\downarrow$ & \textbf{Validity} (\%) $\uparrow$ & \multicolumn{2}{c}{\textbf{Sparsity}} (\%) $\downarrow$ & \textbf{Proximity} $\downarrow$\\
     & & \textbf{Cat} & \textbf{Num} & \textbf{Num} & & \textbf{Cat} & \textbf{Num} & \textbf{Num}\\

    \midrule
    \multicolumn{5}{c}{\cellcolor{black!15}\adult} & \multicolumn{4}{c}{\cellcolor{black!15}\bank}\\
    \midrule
     \wachter & 0.948 & - & 1.000 & 0.778 & 0.803 & - & 1.000 & 1.953\\
     \dice & \textbf{1.000} & \textbf{0.011} & 0.771 & 2.643 & \textbf{1.000} & \textbf{0.164} & \underline{0.822} & 1.347\\
     \revise & 0.879 & \underline{0.196} & 0.993 & 0.965 & 0.418 & \underline{0.177} & 0.945 & 1.032\\
     \cchvae & 0.784 & 0.213 & 1.000 & \underline{0.685} & \underline{0.934} & 0.216 & 1.000 & \underline{0.591}\\
     \tabcf & \textbf{1.000} & 0.402 & \underline{0.513} & 1.106 & \textbf{1.000} & 0.400 & 0.961 & 2.955\\
     \dicoflex & \textbf{1.000} & 0.457 & 0.876 & 1.019 & \textbf{1.000} & 0.317 & 0.908 & 1.081\\
     \textbf{\us} & \underline{0.986} & 0.489 & \textbf{0.250} & \textbf{0.048} & 0.888 & 0.399 & \textbf{0.164} & \textbf{0.052}\\
    \midrule
    \multicolumn{5}{c}{\cellcolor{black!15}\default} & \multicolumn{4}{c}{\cellcolor{black!15}\lending}\\
    \midrule
     \wachter & 0.961 & - & 1.000 & 1.654 & 0.830 & - & 1.000 & 0.771\\
     \dice & \textbf{1.000} & \textbf{0.177} & \underline{0.902} & \underline{0.111} & \textbf{1.000} & 0.230 & \underline{0.861} & 0.558\\
     \revise & 0.221 & \underline{0.225} & 1.000 & 0.226 & 0.783 & \textbf{0.092} & 1.000 & 0.646\\
     \cchvae & 0.901 & 0.239 & 1.000 & 0.505 & 0.842 & \underline{0.118} & 1.000 & \underline{0.502}\\
     \tabcf & \underline{0.999} & 0.466 & 0.999 & 0.717 & \textbf{1.000} & 0.565 & 0.915 & 0.692\\
     \dicoflex & \textbf{1.000} & 0.437 & 0.972 & 0.582 & \textbf{1.000} & 0.434 & 0.961 & 0.874\\
     \textbf{\us} & 0.841 & 0.278 & \textbf{0.214} & \textbf{0.050} & \underline{0.970} & 0.410 & \textbf{0.500} & \textbf{0.268}\\
    \midrule
    \multicolumn{5}{c}{\cellcolor{black!15}\magic} & \multicolumn{4}{c}{\cellcolor{black!15}\htru}\\
    \midrule
     \wachter & 0.658 &  & 1.000 & 0.536 & 0.977 &  & 1.000 & 0.987\\
     \dice & \textbf{1.000} &  & \underline{0.999} & \underline{0.363} & \textbf{1.000} &  & 1.000 & 0.716\\
     \revise & \textbf{1.000} &  & 1.000 & 0.775 & \underline{0.999} &  & 1.000 & 1.269\\
     \cchvae & \textbf{1.000} &  & 1.000 & 0.865 & 0.940 &  & 1.000 & 1.430\\
     \tabcf & 0.943 &  & 1.000 & 0.883 & 0.615 &  & 1.000 & \underline{0.634}\\
     \dicoflex & \textbf{1.000} &  & 1.000 & 0.593 & \textbf{1.000} &  & 1.000 & 0.791\\
     \textbf{\us} & \underline{0.950} &  & \textbf{0.190} & \textbf{0.358} & 0.997 &  & \textbf{0.406} & \textbf{0.580}\\
  \bottomrule
  \bottomrule
  \end{tabular}
  \endgroup
\end{table}

%% file: sections/conclusion.tex



\section{Conclusion and future work}
\label{sec:conclusion}
This paper presents \us, a method that frames counterfactual generation as a sparse transport map for mixed-type data. We solve the problem using flow matching between undesired and desired distributions, which gives a straight displacement vector towards the target class. We use this geometry to optimise for the sparsity and proximity desiderata: a gate restricts displacement to a few features per instance, and the flow is tuned through the gate to improve proximity. Our method \us\ is model agnostic, and across six tabular datasets and six competing methods we produce the sparsest and most proximal numerical edits in one forward pass without querying the black box, and remain comparable on the other desiderata to competitors that assume far greater access (e.g. gradients, logits, etc.) to the model they explain. 


In future work we want to address further CF desiderata, in particular group and global counterfactuals \cite{ley2023globe,rawal2020beyond}, which explain a set of instances rather than a single one. Our transport suits this well: changing the pairing of Step 1 from one-to-many to many-to-many would learn the flow at the group level, and the resulting displacement vectors would be shared across instances by construction. A second direction is interpretability: distilling the learned map into an explainable policy \cite{de2023synthesizing} would state not only what to change but why. 